\documentclass[sigconf,authorversion,nonacm]{acmart}
\usepackage{tikz}
\usetikzlibrary{positioning,arrows.meta,calc,fit,shadows,automata,shapes, calc, arrows, backgrounds}
\AtBeginDocument{%
  }

\begin{document}
%\raggedbottom
%\title{From Simulation to Design Time: Evolution of Influence Models in Cyber-Physical Systems Development}
\title{Towards Enriching Design-Time Models with Evidence from Simulation and Operational Activities}

% \author{anonymous} %\orcidlink{0009-0007-8455-9627}}
% \email{anonymous@anonymous}
% \affiliation{%
%   \institution{anonymous}
%   \city{anonymous}
%   \country{anonymous}
% }

\author{Barbara da Silva Oliveira}
\email{barbara.da-silva-oliveira@univ-cotedazur.fr}

\affiliation{%
  \institution{Université Côte d’Azur,\\ I3S/INRIA Kairos}
  \city{Sophia Antipolis}
  \country{France}
}

\author{Julien Deantoni}% 
  \email{ julien.deantoni@univ-cotedazur.fr}
% \orcidlink{0000-0001-6962-7846} }
\affiliation{%
  \institution{Université Côte d’Azur,\\ I3S/INRIA Kairos}
  \city{Sophia Antipolis}
  \country{France}
}

  \author{Nicolas Ferry }%\orcidlink{0000-0003-2036-0508}}
\email{nicolas.ferry@univ-cotedazur.fr}
\affiliation{%
  \institution{Université Côte d’Azur,\\ I3S/INRIA Kairos}
  \city{Sophia Antipolis}
  \country{France}
}

%\renewcommand{\shortauthors}{da Silva Oliveira et al.}

\begin{abstract}
Cyber-Physical Systems are developed through multiple model-based activities, comprising several stages of development.
%As the development progresses from design to simulation and operation,  
%which can change as the development process progresses from design to simulation and operation,
As the development progresses through simulation, experimentation, and operation, observations of system execution may reveal missing, uncertain, or invalid design knowledge. The path from execution results to persistent design-time enrichment, however, is often implicit. This paper addresses this challenge by proposing an evidence-driven workflow for systematically transferring knowledge produced during simulation and operation back into design-time models. We demonstrate the workflow using an Influence Model, a design-time artefact used to represent how design artefacts and environmental factors jointly affect system response properties, using a sign-following robot use case. %As one of their modelling principles, they evolve iteratively as knowledge about the system’s behaviour grows.
We show how simulation evidence can enrich Influence Models by supporting participant pruning decisions, interaction characterisation, influence-function abstraction, and validity-scope information. The application demonstrates the feasibility of the workflow for simulation-based evidence and its potential benefits to improve design-time reasoning and feedback in subsequent development iterations.
\end{abstract}

%    <concept>
%        <concept_id>10010147.10010341.10010342.10010343</concept_id>
%        <concept_desc>Computing methodologies~Modeling methodologies</concept_desc>
%        <concept_significance>500</concept_significance>
%        </concept>
%    <concept>
%        <concept_id>10010520.10010553</concept_id>
%        <concept_desc>Computer systems organization~Embedded and cyber-physical systems</concept_desc>
%        <concept_significance>500</concept_significance>
%        </concept>
%  </ccs2012>

% \ccsdesc[500]{Computing methodologies~Modeling methodologies}
% \ccsdesc[500]{Computer systems organization~Embedded and cyber-physical systems}


\maketitle

\section{Introduction}

Cyber-Physical Systems (CPS) operate by integrating computational and physical processes, utilizing sensors, actuators, and networked control mechanisms to monitor and adapt to their environments \cite{rajkumar_cyber-physical_2010}. CPS development involves many interrelated artefacts addressing different system concerns, requiring coordination by teams with diverse expertise \cite{griffor_framework_2017}. These artefacts describe the system across various abstraction levels and development stages, capturing different dimensions of system behaviour~\cite{combemale:hal-00994551}.

As CPS development progresses from design to operation, developers continuously acquire new knowledge about the system \cite{madni_model-based_2018}. Simulation campaigns, controlled experiments, and operational monitoring generate vast amounts of data. From these data, empirical evidence can be derived regarding how the system behaves under varying input configurations, design alternatives, and environmental constraints (\textit{e.g.}, statistical summaries or sensitivity-analysis results). Currently, developers must typically analyse this data manually to extract meaningful evidence that can then be used to guide design decisions. This step is critical as CPS behaviour frequently emerges from complex interactions between software, physical dynamics, and the environment, which are rarely fully understood at design time. %However, without a formal methodology, this manual analysis is time-consuming, error-prone, and difficult to trace.

While existing approaches in the domains of Models@run.time and digital twins link simulation or operational data to existing models, their primary goals include runtime synchronization, monitoring, prediction, calibration, or operational decision support \cite{bencomo_modelsruntime_2019,zech_model-based_2026}. %typically data association or monitoring (\textit{e.g.}, performance or safety monitoring).
What is currently missing is a clear process detailing how operational and simulation data can be extracted, analysed, and ingested to generate actionable knowledge (\textit{i.e.}, evidence) that fundamentally enriches Model-Driven Engineering (MDE) models. By enrichment, we refer to the addition of domain-specific information that is directly linked to the model's domain and can be used for design-time reasoning (either by a reasoning engine or a human stakeholder).

To address this gap, we present and then generalize an experience report on an empirical evidence-driven, MDE design model enrichment workflow. This workflow formalizes the step-by-step process of analysing simulation and operational data to derive empirical evidence, and systematically organizing how evidence is interpreted and transformed into enrichment proposals. From a model-management perspective, the workflow governs the creation and application of evidence-backed model deltas. It requires explicit links among evidence items, target model elements, interpretation rationale, validity scope, review decisions, validation results, and repository versions. In this way, the workflow treats enrichment not merely as data analysis, but as a traceable model-evolution activity.

%This workflow treats simulation and operational data as knowledge that can be leveraged to obtain empirical evidence, which in turn must be systematically processed before it can be formalized as model evolution knowledge. 
We instantiate this workflow using \emph{Influence Models} \cite{modelsward26}. An Influence Model is a design-time artefact that represents how various participants, such as system components, environmental factors, and input responses, affect an output system response constrained by requirements. In our experiments, an Influence Model serves dual purposes: as a source to drive simulation campaigns and as a target that, once enriched with empirical evidences, can be used by a stakeholder or a reasoning engine to take informed decisions about how best to control the response of the system under specific environmental factors. 
%once enriched by ingested empirical evidence, becomes analysable by an existing reasoning engine to support stakeholder decision-making.

The paper makes the following contributions:

\begin{itemize}
\item We propose an evidence-driven workflow that makes explicit the path from raw observations from simulation and operation to enriched design-time MDE models.
\item We demonstrate the workflow's steps through experiments on \emph{Influence Models} in the context of a sign-following robot use case, showing how empirical evidence supports design-time reasoning and continuous model evolution during development.
\end{itemize}

The remainder of this paper is organised as follows. Section \ref{sec:background} presents the background. Section \ref{sec:workflow} presents the evidence-driven design-time model enrichment workflow. Section \ref{sec:influence-model} introduces Influence Models as the target design model used to illustrate the workflow. Section \ref{sec:related-work} presents related work. Section \ref{sec:conclusion} concludes the paper and outlines future work.

\section{BACKGROUND AND PROBLEM STATEMENT}\label{sec:background}
% \subsection{Design-Time Models and Execution Evidence in CPS}
% \sebsection{Feedback loops in Existing Work}
% \subsection{}
The development of Cyber-Physical Systems involves multiple engineering phases, including activities such as requirements elicitation, design modelling, simulation, validation, deployment, operation, and maintenance. In each of those phases, several stakeholders, from different domains of expertise, create and maintain heterogeneous artefacts that capture distinct system concerns, assumptions, and views of behaviour.

The complexity of the system can be managed using Model-Based Systems Engineering approaches. Model-Based Systems Engineering (MBSE) is a systems engineering paradigm in which models are treated as primary artefacts throughout the development life cycle \cite{ramos_model-based_2012}. MBSE supports system specification, architecture definition, behavioural description, requirements engineering, verification, and validation \cite{Madni2018Modelbased}. 
In CPS development, MBSE is particularly important because it allows engineers to represent different concerns at different abstraction levels. Architectural models may describe the system structure and the relations between components. Behavioural models may describe modes, transitions, interactions, and execution scenarios. Requirements models may capture expected properties and constraints.

In an MBSE approach, CPS engineering relies on \emph{design models}, which are models used during design-time reasoning, including models of requirements, assumptions, constraints, components, behaviours, relations, architectures, or validity domains. These models are defined during the design phase and are used for subsequent phases. For example, executable simulation models are created to validate system behaviour against what was expected from the design phase.

Design models remain useful only if they remain aligned with the current knowledge available about the system. However, CPS knowledge evolves continuously. As development progresses from design to simulation, experimentation, and operation, engineers observe system behaviour under different configurations and environmental conditions.

The knowledge acquired from the simulation, experimentation, and operation activities then needs to be incorporated into persistent design-time models. In this work, we consider a particular form of \emph{model evolution} denoted as \emph{model enrichment}. We refer to \emph{model enrichment} as the evolution in which \emph{evidence} obtained during execution is interpreted and incorporated into a design-time model to address a declared \emph{domain-specific concern}. Evidence thus helps \emph{refine} the precision or adequacy of the representation of knowledge and may enrich the model structurally or semantically, for example, by modifying or refining model elements and relationships, properties, meanings, or behavioural knowledge.
%domain knowledge more adequate for a declared \emph{domain-specific concern}, and it is then used to \emph{refine} design-time models. 
%In model enrichment, we refer as \emph{refinement} the improvement in the precision or adequacy of the representation of knowledge. %In this sense, model enrichment may include the addition of missing knowledge, removal of unsupported model elements, or revision of relationships. % domain-specific knowledge in a model, so that this knowledge becomes explicit for reasoning.

Model enrichment and model consistency are distinct concepts. Model consistency is defined according to a specified relation over one or more models, and it is triggered when an existing consistency relation has been violated \cite{farber2025observable}. % For example, Färber et al. use observable semantics to characterise heterogeneous models as consistent when the sets of values for a shared observable have a non-empty intersection \cite{farber2025observable}. 
In contrast, model enrichment is triggered by the inadequacy of the knowledge represented by a model for addressing a concern. %In this sense, an enrichment may also add, remove, or revise knowledge in models that are already consistent.

However, \emph{raw observations} from simulation and operation (e.g., traces, logs, sensor values, output values, metrics) do not translate automatically into design-time model knowledge. %They must first be processed into \emph{evidence}. 
In this work, \emph{evidence} refers to information selected or derived from execution results because it is relevant to the \emph{concern}. Evidence can include selected or processed observations, including aggregated output values, and analysis outputs, such as statistical summaries. In other words, a raw observation or an analysis result serves as evidence when it is selected as relevant to an explicit concern within a stated validity scope. Reasoning engines or developers must then interpret the evidence, which results in \emph{interpretation claims} with domain-specific meaning attributed to the evidence. The interpretation claim is then used to formulate an enrichment proposal. %who make decisions on how the design-time models should be modified.

\begin{figure}
    \centering
    \includegraphics[width=0.85\linewidth]{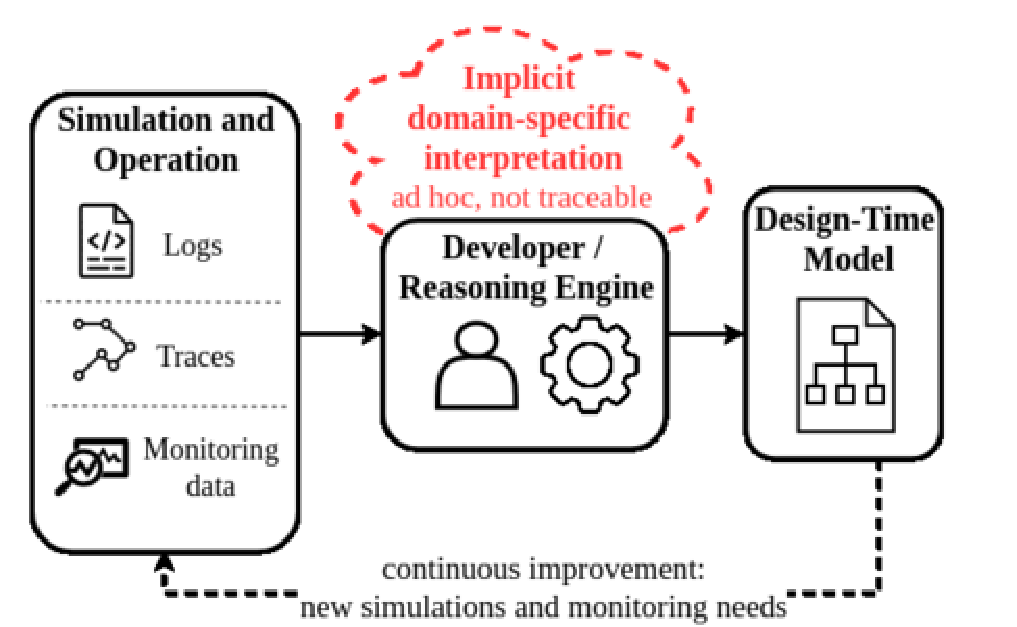}
    \caption{ Domain-specific interpretation to propose design-time model enrichment often remains implicit.}
    \label{fig:challenge}
    \Description[<Domain-specific interpretation to propose design-time model enrichment often remains implicit>]{<Image illustrating the transition from raw observations to evidence and from evidence to model enrichment is often left implicit >}
\end{figure}

The problem addressed in this work is that, in current practice, the transition from raw observations to evidence and from evidence to model enrichment is often left implicit, as illustrated in Figure \ref{fig:challenge}. The numerous steps between producing raw results and enriching models remain unclear, and the domain-specific interpretation often relies on ad hoc decisions.

Horsch et al. identify an epistemic gap in computational engineering information systems, documenting only the produced data is insufficient for data management \cite{horsch2023epistemic}. Data alone are insufficient unless their meaning, the grounds to accept that meaning, and assessments of its validity are also explicit and documented. The challenge in evolving cyber-physical systems is similar, where design-time models may be modified, but the rationale, evidence, validity scope, and domain-specific meaning of the enrichment are not systematically represented in the model.
%As a result, the design-time model may be modified, but the rationale, evidence, validity scope, and domain-specific meaning of the enrichment are not systematically represented in the model.

%, not being systematically formalized, traceable or reusable as part of the model.

%As a result, the models may become incomplete and disconnected from what has been learned in later phases of the development process \cite{boschert_digital_2016}.
% As a result, there is a disconnection between what has been observed and what is represented in the design-time models. 
%Therefore, this data needs to be analysed and interpreted, so it becomes \textit{evidence}, i.e., the knowledge on relevant aspects of system behaviour for engineering purposes, supporting model updates.

% Not relevant. When done manually and without a systematic methodology, this process may require effort, and it is error-prone. Moreover, manual updates hinder traceability, making it difficult to determine why a model was changed, which evidence supported the change, and under what conditions the modification is valid.

This gap motivates the need for a workflow that explains \textbf{how evidence produced and collected during simulation and operation can be transformed into design-time model enrichments}.

\section{EVIDENCE-DRIVEN DESIGN-TIME MODEL ENRICHMENT WORKFLOW}\label{sec:workflow}
This section presents an empirical evidence-driven workflow for model enrichment. The workflow details the required steps to collect raw observations of system behaviour, structure and interpret them, and map the resulting evidence to design-time model elements as enrichment proposals. %Automation may support some steps of the workflow, such as evidence collection, structuring, alignment, and proposal generation, but human review is still required for design-time model modifications.

%The workflow assumes that the target design model has explicit model elements that can be traced to evidences, a method to align evidence variables with model concepts, specific model updates that can be represented explicitly, and validation rules that can be checked before committing the evolved model to a repository.

\subsection{Design Principles}
Following the discussion in Section \ref{sec:background}, we identified principles that guide the workflow. First, %developers identify what model information is missing or unclear, and this drives the selection and extraction of evidence. 
evidence production must start from an explicit \emph{domain-specific concern}. Second, simulation and operation provide different observation contexts, requiring different mechanisms throughout the workflow. Third, evidence must be aligned with target elements and transformed into reviewable model modifications.

%As discussed in section \ref{sec:background}, when engineering CPS, design decisions are hard to be taken  directly from raw logs, traces or output values, it requires interpretation of structured evidence. %In runtime-model research, data acquired from the system and its environment needs to be mediated by models that support monitoring and analysis. In digital-twin work, the real-time data is combined with simulation models and liked to thresholds, risk estimates. 

The workflow starts from an explicit need for model enrichment. Developers first specify what model information is missing or unclear, which model elements are relevant, and which requirement or design question is under consideration. This drives the selection and extraction of evidence. These elements should be defined before evidence acquisition to help ensure that the workflow focuses only on information relevant to the enrichment concern.

%In our workflow, this motivates making explicit the design concern, evidence plan, and the interpretation plan before the proposal of enrichment.

Moreover, simulation and operational data provide different forms of evidence, and, because of that, they need to be treated in different ways. Simulation and experimentation allow developers to explore alternative configurations and understand system behaviour according to design alternatives. In contrast, operational monitoring provides observations of the actual system under real conditions, showing whether the assumptions in the model remain valid. Both activities produce relevant knowledge that can be used for model enrichment. Therefore, the workflow should consider the specificities of each evidence source and may use simulation evidence, operational evidence, or a combination of both.

% Another point is that system observation must be structured before it becomes evidence and it can support action. Raw simulation or operation data needs not only to be filtered, aggregated, but also to be selected and extracted as evidence, so it can be considered useful for model enrichment. As part of the methodology, this should be defined in the stages of definition of evidence and interpretation plans, before the execution of the system during simulation or operation.

%REMOVED BECAUSE OF COMMENT A further point is that the evidence is only valid when the validity context of the model is explicitly captured \cite{van_acker_validity_2024}. This is important because the goal is to provide relevant and meaningful information about the system behaviour, so it can be used for later reasoning. The evidence needs to support not only what should be changed in the model, but also why this change is justified, and what scope of validity should be associated with the enrichment.

Finally, execution and design-time models are expressed using different concepts. Execution data is usually expressed in traces, sensor values, and events, while design models are expressed in terms of requirements, functions, components, states, assumptions, or constraints. Therefore, it is essential to align the evidence with the design elements related to the concern. Moreover, keeping model consistency after enrichment is essential, and this should follow the validation rules of the target modelling language.

\subsection{Workflow description}
\begin{figure*}
    \centering
    \includegraphics[width=0.9\linewidth]{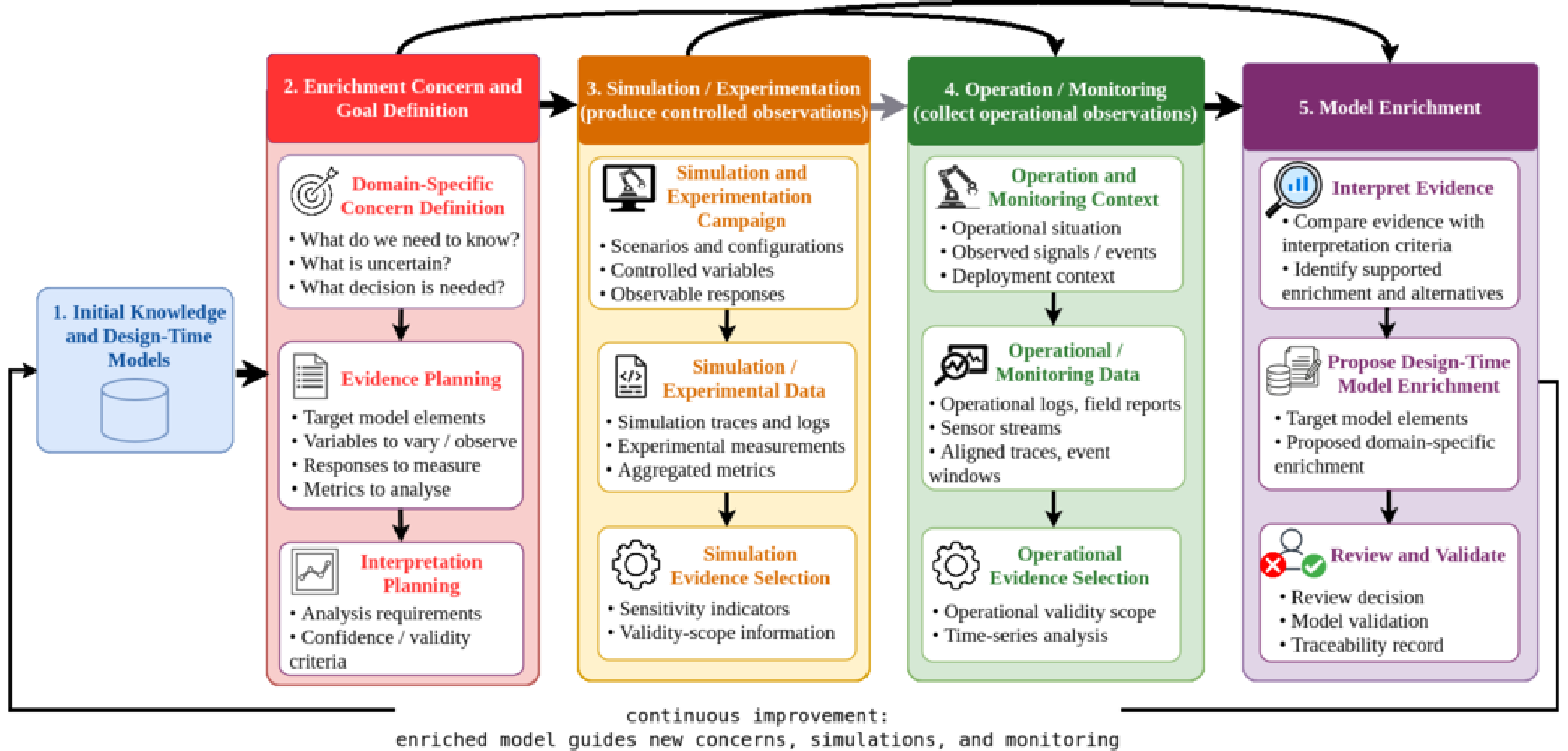}
    \caption{Evidence-driven workflow for enriching design-time models.}
    \label{fig:workflow}
    \Description[<Evidence-driven workflow for enriching design-time models>]{<Workflow comprising the stages of enrichment concern and goal definition, simulation and operation activities, and model enrichment phases. >}
\end{figure*}

% The workflow starts with the definition of the \textit{design model purpose}. This phase includes the identification the target model, definition of what the this model represents, its responsible stakeholders, and its validity scope. This ensures that acquired evidence is assessed only with respect to the model’s intended abstraction and engineering role.
Figure \ref{fig:workflow} presents the proposed evidence-driven workflow for enriching design-time models. The workflow starts with the existing design-time models and the knowledge currently available about the system. These artefacts are the baseline from which stakeholders define a domain-specific concern and specify the intended enrichment. The workflow continues with simulation or operation activities, during which it progressively transforms raw observations into extracted evidence. The evidence is interpreted according to the predefined interpretation analysis, linked to target model elements, and used to formulate model enrichment proposals that are reviewed and, if accepted, committed. It is an iterative process in which the enriched model supports reasoning activities, plans new simulations, guides operation, and produces additional evidence for future enrichment cycles.

\subsubsection{Enrichment Concern and Goal Definition}
The workflow begins when a stakeholder identifies a decision-relevant question that cannot be explained by the current design. This may take the form of knowledge represented in the model that is missing, uncertain, not sufficiently precise, or no longer valid in the model. For example, the stakeholder may need to determine whether an assumption about system behaviour is supported by observations collected during operation or whether a modelled relationship is adequately characterised.

%A concern expresses a decision-relevant limitation, uncertainty, or knowledge need associated with the current design-time model. It precedes evidence acquisition and is operationalised through questions that determine what observations and evidence are required. A concern does not prescribe a model modification and may ultimately be resolved without changing the model.
%By contrast, Cederbladh et al. associate intent with a transactional delta, that is, a group of model operations performed for a common reason. Intent explains why particular modifications were made and relates them to applicable modelling guidelines or engineering standards, thereby supporting review and auditing. Intent is consequently change-centred, whereas concern is problem- and knowledge-centred.

% An \emph{enrichment concern} identifies model knowledge that is missing, uncertain, insufficiently precise, or no longer valid, together with the design-time reasoning task that this knowledge should support. The \emph{enrichment goal} specifies what domain-specific knowledge should be added, refined, removed, or annotated if the evidence supports it.

\begin{figure}
    \centering
    \includegraphics[width=\linewidth]{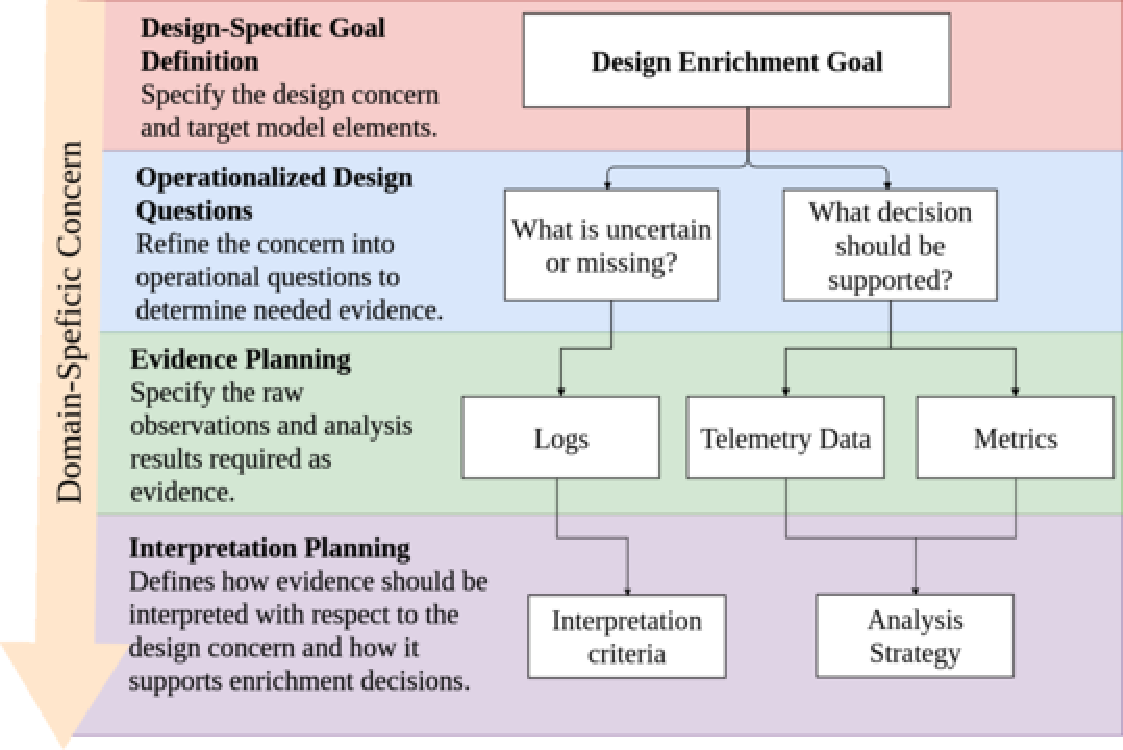}
    \caption{GQM-inspired definition of an enrichment goal and the associated evidence and interpretation plans.}
    \label{fig:gqm-design-purpose}
     \Description[<GQM-inspired definition of an enrichment goal and the associated evidence and interpretation plans>]{<. >}

\end{figure}

This stage will drive the subsequent phases, leading to a model \emph{enrichment} for a specific goal and domain. For this step, we propose a methodology inspired by Goal-Question-Metrics (GQM) \cite{caldiera1994goal}. Figure \ref{fig:gqm-design-purpose} illustrates the application of the process. The developers start from a \emph{design enrichment goal}, derive \textit{operational design questions} from it and specify which specific \textit{evidence} must be collected. The \emph{evidence planning} specifies which information is required to answer each operational design question and produces the evidence plan. This plan identifies the raw observations to be collected, such as logs, filtered traces, aggregated output values, or event windows, as well as analysis outcomes, such as aggregated metrics, indicators or summaries directly derived from execution results. Next, in the \emph{interpretation planning} phase, they provide an \textit{interpretation plan}, describing how the evidence will be evaluated according to the operational design questions to formulate interpretation claims. This plan may comprise analysis strategies, interpretation criteria, decision rules, and validity criteria. The resulting interpretation claims support and justify the formulation of proposals of model enrichment.

%The stakeholder defines the \textit{enrichment goal}, what domain-specific knowledge he expects to add to the model if the evidence collected during simulation or operation activities support it. 

\subsubsection{Simulation/ Experimentation and Operation / Monitoring}
Once the enrichment concern and the evidence and interpretation plans have been defined, the workflow continues with \textit{production or collection of raw observations} through simulation, experimentation, operation or monitoring. Those phases may be used independently or both may contribute evidence related to the same enrichment concern.

The \textit{Simulation/Experimentation} output data may include simulation traces, output results, input configurations, scenario descriptions, and recorded assumptions. The acquisition can be done after each simulation campaign, after a specific experiment, or when an unexpected behaviour is detected during simulation. Those observations are processed according to the evidence planning, which may include estimating uncertainty, computing statistical summa\-ries, or applying analysis methods such as sensitivity analysis. For example, in process mining, it is common to apply techniques for clustering logs, grouping and transforming them before extracting process knowledge \cite{van_der_aalst_process_2016}. Similarly, trace clustering can be used to organize heterogeneous executions into subsets before further analysis.

In the case of \textit{Operation/Monitoring}, observations may be acquired in different ways. In continuous operation, they may include traces and sensor events collected as streams. In this case, acquisition may occur in periodic time or be event-driven, for example, when runtime monitoring detects anomalies or performance antipatterns. In cases of intermittent or batch-based observation, observations are collected from retrieved log batches and monitoring reports. The evidence plan may require operational observations to be processed by online filtering, event-window construction, alignment of traces, or the computation of monitoring indicators.

% Nice but not relevant hereThese two operational modes require different treatment. Continuous streams require message queues or streaming middleware, services for filtering and summarizing large volumes of data, before the use of the collected data. Interrupted operation raises other challenges, for example, the reconstruction or replay of the operational context in which the observations were produced. The evidence plan may require online filtering, time synchronization, event-window construction, or the computation of monitoring indicators.

In Digital Twins approaches, the process may also involve computing metrics from operational measurements and the simulation outputs, since both operation and simulation contribute to assessing the representativeness of models. This requires relating data collected from the physical system to data produced by the simulation model under comparable scenarios, configurations, or assumptions. For example, Mertens and Denil use techniques to compare real-world and simulated data and analyse when a simulation-based digital twin no longer represents the physical system \cite{mertens_reusing_2025}. The resulting deviation or validation metrics may constitute evidence supporting model enrichment.

After either or both stages, the available observations and analysis outputs are selected and structured as \textit{evidence}. %In this workflow, evidence is not all data. 
Evidence consists of data selected and structured according to the domain-specific concern. This stage includes not only cleaning invalid runs and aggregating executions, but also extracting metrics, clustering traces, and computing analysis results such as sensitivity indices.

\subsubsection{Model Enrichment}

The workflow then proceeds to the \textit{Model Enrichment} stage. The workflow \textit{interprets the evidence}, according to the analysis strategy and the validity and confidence criteria defined in the interpretation plan. % Evidence assessment is important, but let's keep out of it, as it is not the focus of this work. The goal is to evaluate how reliable the collected evidence is. This include calculating metrics such as confidence and likelihood estimation, characterizing uncertainty, as well as identifying the validity domain. Evidence may only be meaningful under assumptions and limits of the simulator or experiment ggn. A behavior observed in one simulation campaign or runtime may be valid for a restricted region of the input space, or a specific operational context. Moreover, a runtime event may result from a sensor error, or deployment-specific context instead from misconceptions in design that must be updated.

For each domain-specific concern, the workflow formulates interpretation claims. These claims will support proposals of model-enrichment to the target model. %When evidence is considered relevant for model evolution, the workflow generates one or more \emph{model-update proposals}. 
%This goes beyond monitoring and calibration. The goal is to produce model enrichment proposals that are justified by empirical evidence and remain usable for later design-time reasoning.
An enrichment proposal describes aspects such as the target model elements, the proposed domain-specific enrichment, its supporting evidence, and the interpretation rationale. The proposal may also describe the impact of the enrichment and its validity domain if applicable. Moreover, it may also include alternative interpretations when different model changes could explain the same evidence.

Depending on the type of design model, proposals may involve different types of model modifications. They may suggest, for example, modifying a parameter, adding or removing a relation, pruning a model element, restricting a validity domain, changing a constraint, or adding a traceability link. Each model-modification action must be explicitly associated with its evidence, so that it is actionable and comprehensible to stakeholders.

Once the model enrichment is proposed, it can then be reviewed. Human review governs whether an evidence-supported implication is appropriate for the abstraction, purpose, and engineering intent of the target model. The reviewer may be a developer, domain expert, requirements engineer, or another stakeholder responsible for the design-time model. %The reviewer analyses the proposed enrichment, the evidence that supports it, and take decisions on whether the model enrichment should be accepted. Moreover, stakeholders may propose modification in the proposal before accepting it. %This stage is also important since model modifications may cause conflicts. A change in a model artifact may impact another artifact from another model, which may provoke socio-technical coordination issues. 

Human review is necessary because design-time models encode engineering intent, abstractions, assumptions, and stakeholder decisions. Fully automated workflows are beneficial when rapid response is required, which may be adequate for runtime adaptation or parameter calibration. In this workflow, we consider that it is important to maintain a balanced automation, in which automation may support evidence processing, alignment, and proposal generation, but developers should have control of design-level changes, to better understand the system behaviour and be aware of why the change is proposed and how the supporting knowledge was obtained. This view is consistent with approaches in model repair research. One of the challenges raised in this community is that sometimes several alternative repairs exist, and the selection of the most appropriate one may require developer judgment \cite{macedo_feature-based_2017}.

%The reviewer can approve the proposal as is, accept it with modifications, reject it, or request for more evidence.  If they need more evidence, the process loops back to evidence acquisition. If they suggest modifications, the proposal goes back for generation of enrichment proposal. And if it’s rejected, the decision and the reasons behind it are documented for future reference.

After human approval, the proposed enrichment is applied to the target design-time model. The enrichment is not immediately committed to the design repository. This candidate model is validated to ensure that the proposed modification does not violate model constraints or invalidate related artefacts. This may include syntactic checks, semantic consistency checks, validation of domain-specific constraints, cross-model consistency checks, and behaviour\-al validation. This is aligned with model consistency and continuous engineering approaches, where updates to models must preserve consistency with related artefacts and design assumptions \cite{mazkatli_continuous_2025}.

%If validation fails, the workflow records the violated constraints and the affected artefacts. The proposal is then rejected, and its record may be used to plan future iterations of evidence acquisition. If the new updates or new proposal or models are valid, the enrichment can be applied and committed to the design model repository. %Review this step, should it be send again to developers? ou bien should it be discarted? 

A validated enrichment delta is committed to the design model repository as a versioned model delta. The committed enrichment is related to the supporting evidence, the enrichment proposal, the human review decision, and the validation result. 

The enriched design model then may enable requirement analysis, stakeholder coordination, design-space exploration, the planning of simulation or execution campaigns, and operational guidance. Moreover, the workflow is iterative, as new evidence may lead to the planning of new experiments, and this may lead to additional enrichment proposals. In this way, the design-time model progressively evolves as the knowledge about system behaviour evolves.
    % * Trace Production/Aquisition
    % * Evidence Interpretation
    % * Structure evidence and alignment of Models (which model elements are related with the evidences, semantic mapping )
    % * Evolution/Modification Adaptation Proprosal
    % * Humain Review
    % * Application of changes, creation of new models
    % * Model validation
    % * New Model/Updated Model
\section{WORKFLOW APPLICATION TO INFLUENCE MODELS}\label{sec:influence-model}
To examine the workflow, we have explored how to enrich Influence Models with the evidence produced from simulation. Influence Models were introduced to make explicit relationships among design artefacts, environmental factors, and system response properties in CPS development \cite{modelsward26}. Their purpose is to capture interactions that are often left implicit in standard design models, especially interactions mediated by the physical environment. They are suitable for this application because they can be specified incrementally, from informal descriptions to quantitative functions, and their elements can be enriched as evidence becomes available.

\subsection{Influence Models as Target Design-Time Artefacts}
\begin{figure}[htbp!]
\centering
\usetikzlibrary{shapes.geometric, positioning, arrows.meta, patterns, calc}
\resizebox{\linewidth}{!}{
\begin{tikzpicture}[
    font=\sffamily\large,
    node distance=0.6cm,
    environment_sphere/.style={
        circle,
        minimum size=1cm,
        draw=blue!80!black,
        very thick,
        pattern=horizontal lines,
        pattern color=blue!60!cyan,
        fill=blue!5!white,
        inner sep=0pt
    },
    design_sphere/.style={
        circle,
        minimum size=1cm,
        draw=green!60!black,
        very thick,
        pattern=grid,
        pattern color=green!40!lime,
        fill=green!5!white,
        inner sep=0pt
    },
    SR_sphere/.style={
        circle,
        minimum size=1cm,
        draw=red!60!black,
        very thick,
        pattern=grid,
        pattern color=red!40!lime,
        fill=red!5!white,
        inner sep=0pt
    },
    influence/.style={
        rectangle,
        draw=black,
        thick,
        fill=white,
        minimum width=3.3cm,
        minimum height=1.3cm,
        align=center,
        font=\sffamily
    },
    system_response/.style={
        ellipse,
        draw=purple!80!black,
        thick,
        fill=purple!20!white,
        minimum width=2.5cm,
        minimum height=1cm,
        align=center,
        font=\sffamily
    },
    requirement/.style={
        rectangle,
        draw=black,
        thick,
        dashed,
        fill=white,
        minimum width=2cm,
        minimum height=0.8cm,
        text width=2cm,
        align=center,
        font=\sffamily
    },
    arrow/.style={
        -Stealth,
        thick,
        draw=black
    },
    dashed_arrow/.style={
        -Stealth,
        thick,
        draw=black,
        dashed
    }
]

% Nodes
\coordinate (envcoord) at (0,-0.7);
\node[environment_sphere, pattern color=blue!70!cyan, fill=blue!10] at ($(envcoord.center) + (-0.2, 0.2)$) {};
\node[environment_sphere, pattern color=blue!50!cyan, fill=blue!10!white] at ($(envcoord.center) + (-0.1, 0.1)$) {};
\node[environment_sphere] (envmain) at (envcoord) {};
\node[left=0.25cm of envmain.west, font=\sffamily\bfseries, align=center] (envlabel) {Environmental \\ Factors ($\Phi_{I}$)};

\coordinate (descoord) at (0,-3.5);
\node[design_sphere, pattern color=green!30!lime, fill=green!10!white] at ($(descoord.center) + (-0.2, 0.2)$) {};
\node[design_sphere, pattern color=green!30!lime, fill=green!10!white] at ($(descoord.center) + (-0.1, 0.1)$) {};
\node[design_sphere] (desmain) at (descoord) {};
\node[left=0.25cm of desmain.west, font=\sffamily\bfseries, align=center] (deslabel) {Design Artefacts \\ ($\mathcal{D}_{I}$)};

\coordinate (SRcoord) at (0,-2.1);
\node[SR_sphere, pattern color=red!30!lime, fill=red!10!white] at ($(SRcoord.center) + (-0.2, 0.2)$) {};
\node[SR_sphere, pattern color=red!30!lime, fill=red!10!white] at ($(SRcoord.center) + (-0.1, 0.1)$) {};
\node[SR_sphere] (SRmain) at (SRcoord) {};
\node[left=0.25cm of SRmain.west, font=\sffamily\bfseries, align=center] (SRlabel) {System Response \\ Properties ($\mathcal{SRP}^{in}_{I}$)};

\node[influence, right=of $(envmain.east)!0.5!(desmain.east)$] (inf)
{Influence ($I$) \\ $\mathrm{srp}_I = f(\Phi_{I},\mathcal{SRP}^{in}_{I},\mathcal{D}_{I})$};

\node[system_response, right=of inf] (sysresp)
{System \\ Response\\ Property ($\mathcal{\mathrm{srp}_I}$)};

\node[requirement, above right=0.76cm and -0.34cm of sysresp] (reqbehind1) {};
\node[requirement, above right=0.68cm and -0.42cm of sysresp] (reqbehind2) {};
\node[requirement, above right=0.6cm and -0.5cm of sysresp] (req) {};
\node[above=0.1cm of req.north, font=\sffamily\bfseries] (reqlabel)
{Requirements ($\mathcal{R}$)};

% Arrows
\path[arrow] (envmain.east) -- ($(inf.west)+(0,0.25)$);
\path[arrow] (SRmain.east) -- ($(inf.west)+(0,0)$);
\path[arrow] (desmain.east) -- ($(inf.west)+(0,-0.25)$);
\path[arrow] (inf) -- (sysresp);
\path[dashed_arrow] (sysresp) -- (req);

\end{tikzpicture}
}
\caption{Illustration of the concept of an Influence \cite{modelsward26}.}
\label{fig:influence-model}
    \Description[<Illustration of the concept of an Influence>]{<Illustration of the influence concept, where a set of participants are jointly producing a system response property. >}
\end{figure}
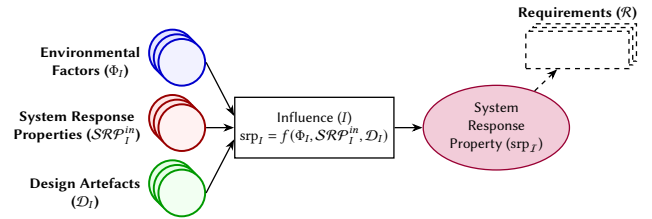

An Influence represents how a set of participants shapes a System Response Property (SRP) \cite{modelsward26}, as represented in Figure \ref{fig:influence-model}. Participants include design artefacts, environmental factors and/or other SRPs. A design artefact is an element created and managed by developers of the system under study, such as parameters, components, or controller settings. An environmental factor is an external element, outside the direct control of developers, such as friction, temperature, noise, terrain, visibility. An SRP is a meaningful behavioural property extracted from the system in operation, such as tracking error, energy consumption, or quality of coverage.

The output SRP may be constrained by a requirement, whose satisfaction is then evaluated against a satisfaction criterion. In this case, an Influence Model provides a traceable path from the coupling of design artefacts and environmental factors to observable system behaviour and requirement satisfaction.

The influence function describes how the participants affect the output SRP. It can be defined at different levels of abstraction, such as natural-language statements, mathematical expressions, rules, or logical predicates. The developers can then express the influence function according to their expertise knowledge, needs and the stage of the development. In some cases, complex equations are not necessary, so the function can be represented in a higher level of abstraction; in other cases, a precise equation is needed. Moreover, at early stages of development, developers may have partial knowledge about environment-mediated behaviour, before the availability of precise equations or empirical data.

An \emph{influence} instance $I$ is a tuple:
\begin{equation}
I \;=\; \bigl(D_I,\; \Phi_I,\; \mathcal{SRP}^{\mathrm{in}}_I,\; \mathrm{srp}_I,\; f_I\bigr),
\end{equation}
where $(D_I)$ denotes the set of design artefacts participating in the influence, $(\Phi_I)$ is the set of environmental factors, $(SRP^{in}_I)$ is the set of input system response properties, $(srp_I)$ is the output system response property, and $(f_I)$ describes how the participants affect the output SRP.

\subsection{Use Case and Initial Influence Model}
The use case is an autonomous sign-following robot based on the TurtleBot3 Waffle Pi platform. The robot operates in an indoor space and navigates a path by detecting visual signs indicating left turns, right turns, and a final stop, realizing the required manoeuvres. The system is developed using domain-specific models and validated through the Simulink/Gazebo co-simulation example provided by MathWorks \cite{MathWorksSignFollowing}. The control and decision logic are implemented in Simulink, while Gazebo emulates the robot and its physical environment through time-synchronized co-simulation.
The robot is evaluated against the requirements:
\begin{itemize}
\item \textbf{RQ1: Sign Search Completion Time.} The time taken by the robot to search for a new sign should be below 4 seconds.
\item \textbf{RQ2: Safe Travel Speed.} The robot must maintain an average speed of at least 0.05 m/s during forward motion, and ensure that its maximum speed does not exceed 1 m/s.
% \item \textbf{RQ3: Reliable Sign Tracking.} The robot must reliably detect and track signs, maintaining a root-mean-square tracking error below 150 pixels.
\end{itemize}

\begin{figure*}
    \centering
    \includegraphics[width=\linewidth]{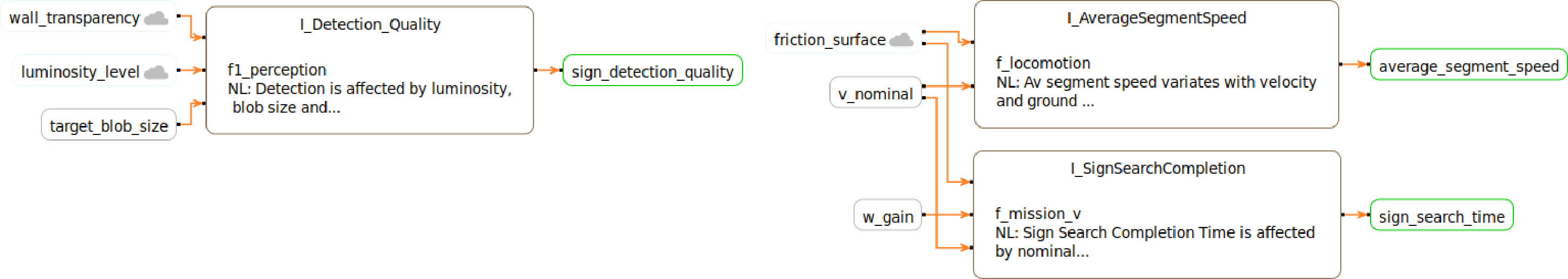}
    \caption{Initial Influence Model for Sign-Following Use Case defined with DemIstifyCPS.}
    \Description[<Initial Influence Model for Sign-Following Use Case defined with DemIstifyCPS>]{<Influence Model comprising influences on detection quality, average segment speed, and sign search completion. >}
    \label{fig:influencesSignFollowing}
\end{figure*}

The application starts from an initial Influence Model that captures partial design-time knowledge about how design artefacts and environmental factors may affect the system response properties. The simplified model comprises three influences, because of space limitations: 
\begin{itemize}
\item \textbf{Sign Detection Quality} 
\(I_{\mathit{DetectionQuality}}\): captures how \texttt{lumi\-nosity\_level}, \texttt{target\_blob\_size}, and \texttt{wall\_transparency} influence the output SRP
\texttt{sign\_detection\_quality}.

\item \textbf{Average Segment Speed} 
\(I_{\mathit{AverageSegmentSpeed}}\): captures how
\texttt{v\_nominal} and \texttt{friction\_surface} influence the
output SRP \texttt{average\_segment\_speed}.

\item \textbf{Sign-Search Completion Time} 
\(I_{\mathit{SignSearchCompletion}}\): captures how
\texttt{v\_nominal}, \texttt{friction\_surface}, and
\texttt{w\_gain} influence the output SRP
\texttt{sign\_search\_time}.
\end{itemize}
Figure \ref{fig:influencesSignFollowing} represents the initial Influence Model. In this first iteration, the Influence Model is not fully specified: developers used natural language to express the influence function and they had low confidence in several definitions, because the definitions were based on prior experience with similar robotic systems, not yet on evidence produced for the current model. %Because of this, the use case is appropriate for exploring how simulation evidence can support the enrichment and validation of design-time Influence Models.

\subsection{Analysing the workflow with Influence Models}

We instantiate the discussed workflow using Influence Models in a sign-following robot use case. Our previous experience designing and conducting simulation experiments for the sign-following robot \cite{da2026conceptual} provided the empirical motivation for the workflow proposed in this paper. Moreover, Influence Models support incremental refinement at multiple levels of abstraction. Their participant sets, function representations, confidence information, and validity scopes can be progressively enriched as evidence is acquired.

The objective of this preliminary application is to show how our workflow supports the systematic transformation of raw observations from simulation into enrichments of a design-time model. The goal is not to provide a full experimental validation of the workflow, but to explore how simulation evidence can be organised, aligned with an Influence Model, and transformed into model enrichment proposals. %We explore if the workflow helps evolving influence models, if it helps clarifying if and which model elements should be evolved, and in a more higher view, if the workflow improves the system development and the understanding of system behavior. %We report which methodological challenges arises at each step of the evolution process.

Starting from an initial influence model, developers then define an evidence plan and an interpretation plan, specifying which simulation data must be produced and how it should be interpreted. %Simulation campaigns produce raw observations, from which evidence will be extracted. 
This evidence is then interpreted to formulate interpretation claims that support model-enrichment proposals. The developers may accept, reject, or edit those proposals, and the approved modifications are applied as model patches to create a new version of the influence model.

\subsubsection{Enrichment Concern and Goal Definition}
During the design-time modelling stage, the developers defined the initial influence models, using the DemIstifyCPS DSL \cite{modelsward26}. These models may be partially defined, incomplete or partially incorrect. For example, the set of participants may be incomplete, and the influence function may be qualitative. These influences serve as the baseline knowledge artefact, and the target for enrichment.

This stage defines the domain-specific concern. The goals of design enrichment are defined according to the needs of developers. In the use case, the main concern is to refine the influence models, so it contains the domain-specific information to support later design-time reasoning. More specifically, the goal is to "Enrich the design-time model so that it provides a more accurate, explicit, and reusable representation of the participants affecting system behaviour, and improve support to design-time reasoning and requirement analysis". % "Can Influence I1 be refined to help understanding the impact of model element X". This concern can then be decomposed into more operational questions: Which participants have a measurable effect on the output SRP? Are some participants negligible within the explored design space? Do participants interact with each other? Can the influence be represented by an analytical function, or is a lookup-table representation more appropriate? Under which input ranges is the resulting interpretation valid? % From there, questions can be raised about the impact of participant, under which operational configurations it has an impact and how to characterize it. The developers then define what would be the logs, telemetric data, or reports that should be taken into consideration as evidence that supports answering the questions.

%In the case of Influence Models, possible enrichments include retaining or pruning participants, updating participant importance, refining the influence function, adding a monotonicity table, updating confidence, or associating the influence with a validity scope. Specifically, in the use case, the operational questions are the following:

This concern is then decomposed into operational questions. For example, developers ask, "which participants should be retained or pruned from each influence?". From these questions, developers determine which type of evidence must be produced to support reasoning about the operational questions, and how it should be interpreted. In this instantiation, evidence is produced through simulation campaigns. %For each influence, the participants define the input space to explore, while the output SRP defines the traces and values to observe. 
%After the simulation, analyses methods are used, such as Morris sensitivity analysis, to verify each influence's participant set against SRP outcomes. %A participant is considered a pruning candidate when its estimated contribution to output variation, approximated by \(\mu^* \cdot \Delta x\), remains below the aleatory uncertainty baseline \(\sigma_{nat}\) within the explored validity scope.

Table \ref{tab:interpretation-plan-influences} summarizes the evidence and interpretation plan used in the sign-following robot use case, providing each operational question, the evidence to collect, and the interpretation criterion.

\begin{table*}
\centering
\small
\caption{Evidence and interpretation plan for enriching Influence Models in the sign-following robot use case.}
\label{tab:interpretation-plan-influences}
\begin{tabular}{p{0.30\linewidth} p{0.3\linewidth} p{0.32\linewidth}}
\toprule
\textbf{Operational question} &
\textbf{Evidence Plan} &
\textbf{Interpretation Criterion} \\
%\textbf{Possible enrichment} \\
\midrule

\textbf{Participant relevance} 
Which participants should be retained or pruned from each influence?
&
Collect simulation runs varying the current participants and observing the output SRP. Compute the Morris sensitivity indices \(\mu^*\), and estimate aleatory uncertainty \(\sigma_{nat}\) from repeated runs.
&
A participant is considered relevant when its estimated effect on the output SRP is greater than aleatory uncertainty within the explored validity scope. 
A participant is proposed for pruning when its estimated effect magnitude remains below the uncertainty estimate.  \\
% &
% Retain participant; prune participant; request additional simulations; annotate the decision with validity scope and supporting evidence. \\
\midrule

\textbf{Participant importance} 
How strongly does each participant affect the output SRP?
&
For each retained participant, compute \(\mu^*\) and derive normalized importance values, for example by normalizing with respect to the maximum \(\mu^*\) or the sum of all \(\mu^*\) values in the influence.
&
Participants with higher normalized \(\mu^*\) values are interpreted as having stronger relative importance for the output SRP within the explored domain. 
The ranking is interpreted only within the corresponding influence and simulation scope. \\
% &
% Add or update participant weights; rank participants; update confidence in participant relevance; preserve importance values as model annotations. \\
\midrule

\textbf{Interaction and non-linearity} 
Do participants interact or produce non-additive effects?
&
Collect joint participant configurations and Morris indices \(\mu^*\) and \(\sigma\).
&
Compare the \(\mu^*\) and \(\sigma\) indices. A high \(\sigma\) value, especially in the case of large relative to \(\mu^*\), suggests non-linear behaviour or interaction effects.  \\
% &
% Annotate interaction indicators; preserve coupling information; guide functional refinement campaigns; avoid overly simple additive influence-function abstractions. \\
\midrule

\textbf{Influence-function abstraction} 
Can the relation between participants and output SRP be represented by an analytical approximation or a monotonicity lookup table?
&
Collect structured datasets relating participant configurations to output SRP values. 
For retained participants, compute local trends, slopes, monotonicity information, and, when appropriate, candidate fitted functions or lookup tables.
&
If the relation is sufficiently regular and the approximation is adequate, an analytical or semi-analytical function may be proposed. 
If the relation is not globally simple but shows local monotonic trends, a monotonicity lookup table should be proposed. \\
% &
% Add analytical influence function; add monotonicity lookup table; add conditional function abstraction; update confidence in \(f_I\); request additional data if the results remains unclear. \\
\midrule

\textbf{Validity scope} 
Under which explored input ranges and simulation assumptions is the enrichment valid?
&
For each campaign, collect participant ranges, fixed baseline values, scenarios, simulator settings, model version, number of repetitions, random seeds, SRP extraction procedure, and uncertainty estimates.
&
An enrichment is valid only for the explored ranges, scenarios, assumptions, and simulation settings that produced the supporting evidence. 
Evidence obtained in one region of the design space must not be generalized to other untested regions without additional simulation or operational evidence. \\
% &
% Attach validity scope; specify enrichment applicability; identify new evidence needs for not explored regions. \\
\bottomrule
\end{tabular}
\end{table*}

\subsubsection{Simulation Campaigns}

Following this, Simulation Campaigns were defined and realised. For each influence, the participants ranges defined the design space to explore in the current campaign, and the output SRP determined the response to be observed.

The simulation campaigns consisted of two coupled campaigns, one focused on understanding the influences structurally and the other on understanding them functionally. The simulation campaigns produced raw observations, including logs, output values, participant configurations, scenario descriptions, and execution metadata. These observations were then analysed based on the evidence and interpretation plans, producing derived outputs such as sensitivity analysis results, interaction indicators, aggregated SRP values, uncertainty estimates and aggregated summaries of simulation runs.

%These artefacts are not directly treated as model knowledge. They are raw observations that must be selected and extracted, and interpreted before they can support enrichment of the Influence Model.

% remove For the structural campaign, the mean absolute elementary effect $(\mu^*)$ is used to estimate participant importance, while $(\sigma)$ indicates the presence of non-linearities or interactions. The aleatory uncertainty $(\sigma_{nat})$, estimated from repeated runs, is used as a baseline to distinguish meaningful effects from natural variations of values.
%remove The functional campaign results captures how the output SRP varies with respect to influential participants over specific ranges of the input space. It produces many representations of the influence function, fitting into regression models when possible, and providing monotonicity lookup tables when the relation is not fitted. 

%\subsubsection{Concern-driven Evidence Selection and Structuring}
The collected simulation results were processed to produce concern-relevant evidence dataset, containing observations and analysis outputs that helped answer operationalized design questions. %As illustration of the structured evidence, the sensitivity indices and uncertainty estimates of the campaigns are represented in Table \ref{tab:structural-sensitivity-evidence}.
Each evidence record was linked to a specific validity-scope, participant configuration, observed SRP, model version, and simulator settings. This traceability is essential since model enrichments are not derived directly from raw logs or CSV files. 

% remove The third stage aligns structured evidence with Influence Model elements. The influences are structured around their definition, where the coupled participants connects to the SRPs, the metadata related to the participants, the influence functions. Each evidence record is linked to a specific influence, campaign, participant configuration, output SRP, model version, simulator settings.

\subsubsection{Model Enrichment}

During the Model Enrichment stage, the selected evidence is evaluated according to the interpretation criteria defined in the interpretation plan. %From the structural campaign, this evidence helps in making decisions to retain, prune, or add participants. For these decisions, the rule is related to the aleatory uncertainty, a participant is considered negligible when its estimated contribution to the output SRP is not distinguishable from the aleatory uncertainty observed in repeated simulations, within the explored validity scope. From the functional campaign, evidence is extracted to enrich the influence function definition. It is possible to add an analytical fit, a monotonicity table, or a participant impact function, or update function confidence. These form a set of candidate model modification operations, which are grouped into a refinement proposal, such as a model patch. The patch represents how the model should evolve, each operation is linked to the respective simulation evidence. 
%In the structural campaign of the use case, sensitivity analysis was interpreted to assess the participant set of each influence. %The results of this analysis clarifies participants who do not significantly influence the SRP, and other elements may be found that has an influence in the system response property. 
In the structural campaign, the Morris sensitivity indices and uncertainty estimates were interpreted to assess the participants of each influence. This helps identify participants that do not significantly influence the SRP when the estimated effects cannot be distinguished from aleatory uncertainty. Such participants are interpreted as pruning candidates. %Furthermore, the second campaign used the sensitivity indices generated by this stage to describe participant importance and their mutual couplings, informing the design of new simulation campaigns.

The second campaign focused on refining the influence function $f_I$ using different function abstractions to better analyse the influence. Possible representations include an analytical expression, a fitted regression model, or a monotonicity lookup tables when no suitable global fit is available. In this instantiation, it produced a monotonicity lookup table. This table records how the output SRP changes across the explored participant ranges and provides a lightweight abstraction of \(f_I\) for later design-time reasoning. 

Table \ref{tab:model-enrichment-results} summarizes the model enrichment proposals according to the domain-specific concern definition. The resulting proposals define candidate model modification operations. Those may be grouped into enrichment patches, describing how the current model should evolve. Each operation should be linked to the respective simulation evidence that justifies it.

\begin{table*}[h!]
\centering
\small
\caption{Candidate model-enrichment proposals derived from the sign-following robot use case according to the evidence and interpretation planning.}
\label{tab:model-enrichment-results}
\begin{tabular}{p{0.16\linewidth} p{0.27\linewidth} p{0.29\linewidth} p{0.20\linewidth}}
\toprule
\textbf{Operational question} &
\textbf{Evidence Obtained} &
\textbf{Evidence Interpretation} &
\textbf{Model-enrichment proposal} \\
\midrule

\textbf{Participant relevance} 
Which participants should be retained or pruned?
&
For \(I_{DetectionQuality}\), Morris analysis produced
\(\mu^*(\texttt{BlobSize}) = 0.2270\),
\(\mu^*(\texttt{AmbientLighting}) = 0.0120\),
and \(\mu^*(\texttt{WallTransparency}) = 0.0071\), with \(\sigma_{nat}=0.0048\).
The estimated contribution of \texttt{WallTransparency} was \(\mu^* \cdot \Delta x = 0.00071\).
&
\texttt{BlobSize} is the dominant participant and \texttt{AmbientLighting} has a measurable effect.
The estimated contribution of \texttt{WallTransparency} is lower than the aleatory uncertainty baseline, meaning that its effect cannot be clearly distinguished from simulation noise within the explored scope.
&
Retain \texttt{BlobSize} and \texttt{AmbientLighting}.
Propose pruning \texttt{WallTransparency} from \(I_{DetectionQuality}\) within the explored validity scope. \\
\midrule

\textbf{Participant importance} 
How strongly does each participant affect the output SRP?
&
For \(I_{\mathit{SignSearchCompletion}}\), Morris analysis produced $\mu^{*}(v_{\mathrm{Nominal}})=1.8000$, $\mu^{*}(\mathit{GroundFriction})=2.4375$, and
\(\mu^*(\mathit{wGain})=2.3063\). Normalizing with respect to the sum of \(\mu^*\), the corresponding relative-importance scores are \(0.2751\), \(0.3725\), and \(0.3524\), respectively.&
%The \(\mu^*\) values provide an estimate of relative participant importance within each influence and explored scope.
Sensitivity indices rank participants by their effect on the output SRP, where $\mathit{GroundFriction}$ has the highest estimated relative importance, followed by $\mathit{wGain}$, and $v_{\mathrm{Nominal}}$.
&
Add or update participant importance annotations.
Use normalized sensitivity indices as participant importance values. \\
\midrule

\textbf{Interaction and non-linearity} 
Do participants interact or produce non-additive effects?
&
For \(I_{AverageSegmentSpeed}\), \(\sigma(\texttt{GroundFriction}) = 0.05659\) and \(\mu^*(\texttt{GroundFriction}) = 0.03102\).
%For \(I_{SignQuality}\), the dispersion values indicate non-additive effects for \texttt{AmbientLighting} and \texttt{WallTransparency}.
&
High dispersion values suggest that the relation should not be treated as purely additive.
For \(I_{AverageSegmentSpeed}\), \texttt{GroundFriction} has a secondary effect, according to its resulting $\mu^*$ but contributes to non-linear or interaction behaviour.
&
Annotate interaction or non-linearity indicators.
Use these indicators to guide functional refinement campaigns, to avoid simplifying influence-function abstractions. \\
\midrule

\textbf{Influence-function abstraction} 
Can \(f_I\) be represented by a lightweight function abstraction?
&
For \(I_{AverageSegmentSpeed}\), the functional campaign produced a monotonicity lookup table.
The local slopes are positive across the explored \texttt{GroundFriction} intervals:
\(0.7803\), \(0.8062\), and \(0.8902\).
&
\texttt{AverageSegmentSpeed} increases with \texttt{vNominal} over the explored domain, and the local slope also depends on \texttt{GroundFriction}.
&
Add a monotonicity lookup table to \(I_{AverageSegmentSpeed}\).
Use it as a lightweight abstraction of \(f_I\) for later design-time reasoning. \\
\midrule

\textbf{Validity scope} 
Under which conditions is the enrichment valid?
&
Each campaign records the explored participant ranges, fixed baseline values, simulation scenarios, simulator settings, model version, repeated runs, and uncertainty estimates.
% For example, \(I_{FloorSlipperiness}\) is analysed over the explored ranges of \texttt{vNominal} and \texttt{GroundFriction}.
&
The enrichments are valid only within the explored simulation ranges and assumptions.
A pruning decision or monotonicity relation should not be generalized to untested regions without additional evidence.
&
Attach validity-scope information to each enrichment.
Preserve traceability links to the campaign, extracted evidence, simulator settings, and uncertainty estimates. \\
\bottomrule
\end{tabular}
\end{table*}

The present instantiation focuses on deriving and documenting these candidate enrichments using simulation evidence in a systematic way. The next activities are not yet fully automated. They are therefore described here as required activities of a complete implementation of the workflow.

Following this, in a complete implementation, the proposal would be submitted to domain-specific experts for review. They could accept the complete proposal, accept only selected operations, modify the proposal, reject it, or request additional evidence. The review decision would then be stored as part of the patch metadata to preserve the history of decisions regarding model evolution.

Once an enrichment proposal is approved, the enrichment patch would be applied to create a candidate enriched model. This candidate would then be validated against the well-formedness constraints of Influence Models, including valid participant definitions, valid numerical ranges, and preservation of the traceability from simulation evidence to the model.

Validated patches would be committed to the model repository as versioned deltas. After the patch application, a complete implementation should also generate an enrichment report. The report documents the differences between the previous and current model versions, the operations that were applied, and the developers' decisions. This allows developers to understand what was changed, why the change was proposed, and why it was accepted.

The enriched model could then be used for design-time reasoning, providing evidence-backed information for later reasoning, such as abstractions of the influence function, interaction effects and participant relevance. This information could improve not only the representativeness of the design-time model itself but also the capabilities of subsequent analyses, allowing developers to better comprehend system behaviour and support discussions of trade-offs among alternative design decisions.

\section{RELATED WORK}
\label{sec:related-work}

The workflow presented in this paper relates to several research areas that study how data, runtime observations, simulation results, or operational knowledge can be connected to models.% However, it is usually only for specific concerns, model types, or lifecycle stages. 

A first relevant line of work is \textit{Models@Run.time}. Runtime-model approaches show that runtime models can remain causally connected to the running system through feedback loops to support monitoring, analysis, and adaptation \cite{vogel_role_2011,bencomo_modelsruntime_2019}. They demonstrate that runtime data can be represented and reasoned about through models. However, these approaches often focus on adapting runtime models, rather than the transfer of evidence into persistent design-time models for future engineering decisions.

Another important area of research concerns \textit{digital twins}. Digital twin research uses runtime data to maintain the representativeness of models. A digital twin can be understood as a virtual representation of a real-world system that evolves through data collected from that system \cite{zech_model-based_2026}. A recent review emphasize physical-to-virtual and virtual-to-physical data flows \cite{thelen_comprehensive_2022}. In \cite{mertens_reusing_2025}, the authors use model validation techniques to detect when the model no longer accurately reflects the physical system, comparing measured and simulated data, and propose calibration or alternative-model search. While those approaches are foundational for maintaining the accuracy of simulation models, most works do not address how such observations can be systematically structured as empirical evidence and how they relate to design-time models.

A recent research in digital twins tackles the data to decision gap. %, where runtime-time data can be combined with simulation models to support what-if analysis, uncertainty estimation, and be linked to decision thresholds \cite{pham_data_2026}. 
In this work, the authors propose a framework to organize sensing systems, analytical models, digital twins, and operational decision processes in a closed-loop architecture \cite{pham_data_2026}. The objective is to transform the infrastructure data into actionable engineering decisions and operational actions, such as maintenance planning, control adjustments, or risk mitigation. This work provides a strong foundation for our workflow, emphasizing that data availability alone does not directly ensure better decisions, and that data, models, and decision mechanisms must be explicitly connected. However, the focus is on supporting operational decision-making and infrastructure management. In contrast, our workflow addresses a complementary gap, where we focus on how evidence from simulation, experimentation, and operation is interpreted with respect to the design concerns, and can enrich persistent MDE design models.

%These approaches are relevant because they show how operational evidence can trigger model evolution. However, they are focused on validating and calibrating simulation models, rather than on design-time model update.

Closely related work concerns \textit{continuous engineering}. The Continuous Integration of architectural Performance Models uses monitoring data, calibration, and self-validation to update architectural performance models in response to changes in operation \cite{mazkatli_continuous_2025}. Such approaches demonstrate that evidence from execution phases can trigger systematic model evolution; however, they are usually specialised for performance models and performance prediction.

\textit{Data assimilation} is another relevant area, focusing on obtaining the best possible description of a dynamical system and proposing the correction of models, typically their numerical states or parameters, accounting for uncertainty \cite{evensen_data_2022}. In \cite{donato_self-updating_2024}, runtime observations of a digital twin are combined with simulation outputs to improve an estimation algorithm. This body of literature is important for our work because it provides a rigorous foundation where system observations are treated as knowledge for model correction; however, they mostly target numerical states, parameters, or simulation models.

%Another important body of work concerns \emph{Validity-Frames Modelling}. Van Acker et al. argue that reasoning about a model is only justified if the validity context of that model is captured explicitly \cite{van_acker_validity_2024}. This is relevant for our workflow because evidence-driven enrichment should not simply propose what to change in a model, but also make explicit why the change is justified and under which assumptions or conditions it remains valid.

\textit{Feedback-driven requirements engineering} offers an important perspective. This line of work investigates how user feedback can support requirements evolution. FAME includes monitoring data as a support for requirement evolution, which helps requirements engineers identify, refine, or update requirements, facilitating design-time modifications based on runtime evidence \cite{oriol_fame_2018}. Such approaches show that runtime evidence can support design-time evolution of engineering artefacts, not only runtime adaptation; however, the primary target in that line of work is requirement evolution.

A further source of inspiration is \textit{process mining} and related log-analysis work. Process mining extracts process knowledge from event logs \cite{marin-castro_event_2021}. Process discovery derives process models from logs, conformance checking compares observed logs with existing process models, and process enhancement extends or improves existing process models using information extracted from logs \cite{van_der_aalst_process_2016}. Work on trace clustering further illustrates how structuring raw logs is used to extract process knowledge, for example by splitting heterogeneous traces into more homogeneous subsets before further analysis \cite{song_trace_2009}. These studies are important for our work as they demonstrate an important methodological point that raw execution data must be cleaned, grouped, filtered, and reorganised before it can support model-level updates.

Another closely related area is \textit{Architecture-Runtime Traceability}. % Arcelli et al \cite{arcelli_exploiting_2019} provide architecture and runtime traceability, explicitly connecting runtime information with design models. 
In this research area, runtime logs are transformed into log models and linked to design artefacts such as UML architectural elements through traceability relationships \cite{arcelli_exploiting_2019}. Runtime performance data can then annotate UML models, support the detection of performance antipatterns and generate refactoring proposals.
This line of work is relevant to our work because it addresses the gap between execution-level observations and design-level artefacts. However, it is often specialized to a particular concern, such as performance, or to a specific modelling language or architectural framework.

%Model Repair! Model repair work generates alternative repair updates for inconsistent models, while recognizing that developers need guidance,

% Model consistency and model repair research provides the understanding of how changing one model element may affect other elements, artefacts, or constraints. 

%Model consistency and co-evolution research shows that changing one model often requires changes in related models. This is important for evidence-driven model update because applying a local change may break constraints, invalidate assumptions, or affect other artefacts.
%Even in the domain of software engineering, there are similar challenges, as in FAME \cite{} discusses how the requirements must be updated according to the feedback and the monitoring of runtime activity, combining those two results and providing proposals for requirement engineers to review, for the new proposal of modifications in requirements. In CPS, multiple approaches in Models@Run.time uses the runtime data to update the runtime models. Here the objective is to improve resource usage, detect and eliminate system anomalies. 

Model-consistency management deals with a closely related but distinct challenge. Färber et al. introduce observable semantics as a way of characterising consistency between heterogeneous models, where each model constrains the admissible values of a shared observable, and consistency holds when these constraints admit at least one common value \cite{farber2025observable}. % Their observables provide semantic interfaces between modelling domains and should not be confused with the execution observations used in our workflow. 
Their work provides a formal basis for specifying and detecting cross-model consistency. Our workflow instead addresses how execution results are selected as concern-relevant evidence, interpreted, and transformed into reviewed model deltas. Enrichment may also improve an already consistent model by making its represented knowledge more adequate for a declared concern.

At the model-evolution level, Cederbladh et al. propose a framework supporting engineering reviews of horizontal model evolution \cite{cederbladh2025traceability}. Their framework represents concrete model modifications as atomic deltas and groups these modifications into intent-annotated deltas. %The intent connects the modifications to the engineering guidelines that motivate them, and it is used in horizontal evolution to derive additional views at the same abstraction level.
Their notion of intent and our notion of concern both make the rationale for model evolution explicit. However, intent is associated with a delta and explains why a set of model modifications was performed. By contrast, an enrichment concern motivates evidence planning, evidence selection, and interpretation.
%The intent has a similar concept as concern, but it is exploited for different purposes. 
%Our concept of concern is at the begninning, and it is what motivates the enrichment, whereas intent is related to the delta for the model modification. %However, Cederbladh et al. focus on recording intent during horizontal evolution to derive additional views at the same abstraction level. Our workflow extends traceability upstream from a domain-specific concern, through execution observations, selected evidence, and interpretation claims, to a reviewed and validated delta.

In summary, existing work investigates important mechanisms related to the loop from execution evidence back to the models, including the production, analysis, and alignment of execution information with models. However, most approaches target a specific artefact or a specific kind of update, such as runtime adaptation, operational decision-making or parameter calibration. They do not make explicit the complete process of transferring knowledge from execution phases to persistent design-time models. %This gap motivated the workflow proposed in this paper.
Our work focuses on clarifying how observations are chosen and extracted as concern-relevant evidence, how evidence can be interpreted according to domain-specific concerns, and how candidate enrichments can be reviewed, validated, and committed to the model repository.
%Across these communities, existing work investigates important stages of the loop from execution evidence back to the models, producing, analysing, and relating execution information to models. %Requirements feedback approaches use runtime logs to support requirements elicitation evolution. Digital twin validation approaches use operational data and validation metrics to calibrate simulation models. Models@Runtime approaches maintain runtime models causally connected to the running system. Process mining extracts process knowledge from event logs. Architecture and runtime traceability approaches align runtime data with architectural model elements. Model-based DevOps approaches maintain model consistency across development and operation. 

\section{CONCLUSION AND FUTURE WORK}\label{sec:conclusion}
This paper proposes an evidence-driven workflow for transferring knowledge produced during simulation and operation back into design-time models. The workflow makes explicit the necessary steps to bridge execution and design-time. These steps includes defining domain-specific concerns, planning and selecting evidence, interpreting the selected evidence, and generating model enrichment proposals.

Raw observations from simulation and operation activities such as logs, traces, and monitoring data do not automatically enrich models. Evidence is selected from the observation based on a domain-specific concern, becoming useful for enrichment only when it carries relevant knowledge related to model elements. Interpretation criteria define how this selected evidence supports a possible model enrichment. This separation makes the reasoning more explicit, ensuring that model modifications do not rely on undocumented or ad hoc decisions.

We applied the workflow to Influence Models, in a sign-following robot use case, and showed how simulation evidence can be systematically transformed into candidate model enrichment operations.

This work is conceptual, and future work will focus on implementing the complete workflow. We plan to evaluate the workflow further using operational and monitoring evidence, additional model types and CPS domains, and to study its use with domain experts. These evaluations should examine not only whether the workflow produces valid model enrichments, but also the whether it improves traceability, model usefulness, decision quality, and developer understanding.

\bibliographystyle{ACM-Reference-Format}
\bibliography{main}

\end{document}